\documentclass[runningheads]{llncs}
\usepackage{amsmath,amssymb,amsfonts}
\usepackage{marvosym}
\usepackage[T1]{fontenc}
\usepackage{multirow}
\usepackage{cite, hyperref, tikz}
\usepackage{orcidlink}
\hypersetup{
            colorlinks=true,
            linkcolor=blue,
            anchorcolor=blue,
            citecolor=blue
            }

\begin{document}

\title{DSSAU-Net:U-Shaped Hybrid Network for Pubic Symphysis and Fetal Head Segmentation}
%
%\titlerunning{.}
% If the paper title is too long for the running head, you can set
% an abbreviated paper title here
%

\author{Zunhui Xia$^{\orcidlink{0009-0008-6706-5817}}$ \and Hongxing Li$^{\orcidlink{0009-0002-7958-3976}}$ \and Libin Lan$^{\orcidlink{0000-0003-4754-813X}}$${(\textrm{\Letter})}$}

% \authorrunning{Z. Zhou}
\authorrunning{Z. Xia et al.}
% First names are abbreviated in the running head.
% If there are more than two authors, 'et al.' is used.
%
\institute{Chongqing University of Technology, Chongqing 400054, China \\
\email{lanlbn@cqut.edu.cn}
}
\maketitle              % typeset the header of the contribution

\begin{abstract}
In the childbirth process, traditional methods involve invasive vaginal examinations, but research has shown that these methods are both subjective and inaccurate. Ultrasound-assisted diagnosis offers an objective yet effective way to assess fetal head position via two key parameters: Angle of Progression (AoP) and Head-Symphysis Distance (HSD), calculated by segmenting the fetal head (FH) and pubic symphysis (PS), which aids clinicians in ensuring a smooth delivery process. Therefore, accurate segmentation of FH and PS is crucial. In this work, we propose a sparse self-attention network architecture with good performance and high computational efficiency, named DSSAU-Net, for the segmentation of FH and PS. Specifically, we stack varying numbers of Dual Sparse Selection Attention (DSSA) blocks at each stage to form a symmetric U-shaped encoder-decoder network architecture. For a given query, DSSA is designed to explicitly perform one sparse token selection at both the region and pixel levels, respectively, which is beneficial for further reducing computational complexity while extracting the most relevant features. To compensate for the information loss during the upsampling process, skip connections with convolutions are designed. Additionally, multiscale feature fusion is employed to enrich the model's global and local information. The performance of DSSAU-Net has been validated using the Intrapartum Ultrasound Grand Challenge (IUGC) 2024 \textit{test set} provided by the organizer in the MICCAI IUGC 2024 competition\footnote{\href{https://codalab.lisn.upsaclay.fr/competitions/18413\#learn\_the\_details}{https://codalab.lisn.upsaclay.fr/competitions/18413\#learn\_the\_details}}, where we win the fourth place on the tasks of classification and segmentation, demonstrating its effectiveness. The codes will be available on \href{https://github.com/XiaZunhui/DSSAU-Net}{GitHub}.

\end{abstract}

\section{Introduction}
Difficult labor is a major cause of maternal mortality and morbidity, referring to situations where, despite strong uterine contractions, parts of the fetus do not pass through the birth canal. 
Therefore, during labor, to prevent such occurrences, the position of the fetus needs to be monitored repeatedly.
Traditional vaginal examinations  are subjective and potentially invasive \cite{boyle2013primary,cohen2017prolonged}, and some related methods are difficult to perform reliably \cite{dupuis2005fetal}. 
The advent of ultrasound-assisted diagnosis has provided a non-invasive, accurate alternative for assessing fetal position and cervical dilation, gradually gaining acceptance in obstetrics \cite{fiorentino2023review}. Some studies \cite{bellussi2017intrapartum,malvasi2014occiput,ramphul2014instrumental,challenge2013evaluation} have shown that intrapartum ultrasound assessment can help more accurately evaluate the positions of the fetal head and the pubic symphysis. The International Society of Ultrasound in Obstetrics and Gynecology recommends ultrasound assessment before considering instrumental vaginal delivery or suspecting delayed labor. Two reliable ultrasound parameters—the angle of progression (AoP) and the head-symphysis distance (HSD)—are used to predict the outcome of instrumental vaginal delivery \cite{ghi2018isuog}.

In ultrasound-assisted diagnosis, the AoP and HSD are key parameters for assessing the progress of labor. These parameters rely on the segmentation of the fetal head (FH) and pubic symphysis (PS), aiding clinicians in real-time monitoring of the fetal delivery status.  Therefore, many methods have begun to focus on improving the segmentation performance of FH and PS. For example, the Fetal Head–Pubic Symphysis Segmentation Network (FH-PSSNet) \cite{chen2024direction} uses an encoder-decoder framework, which incorporates a dual attention module, a multi-scale feature screening module, and a direction guidance block, for automatic AoP measurement. The Dual-path Boundary-guided Residual Network (DBRN)  \cite{chen2024fetal} integrates a multiscale weighted module (MWM), an enhanced boundary module (EBM), and a boundary-guided dual-attention residual module (BDRM) to address the challenges of implementing fully automated and accurate FH-PS segmentation in cases of low contrast or ambiguous anatomical boundaries. 
Additionally, BRAU-Net \cite{cai2023pubic} uses only region-level sparse tokens for FH-PS segmentation, which is not robust enough for small targets due to speckle noise, ultrasound artifacts, and blurred target boundaries.

In contrast to these methods, we consider combining a sparse attention mechanism with convolution and adopting a multiscale feature fusion approach to achieve more effective segmentation of FH and PS. To this end, in this paper, we propose a novel U-shaped network architecture with sparse self-attention, termed DSSAU-Net, for the segmentation of the FH and PS. Specifically, the adopted sparse self-attention, derived from the idea of concentrating tokens at both region and pixel levels \cite{ Zhu2023biformer, Zhao2019ExplicitST}, is a content-aware, dynamic mechanism, which is explicitly designed as dual selection operations and has been used in our other unpublished work. We call this sparse mechanism Dual Sparse Selection Attention (DSSA). DSSA significantly reduces computational complexity while extracting more accurate features. Additionally, to effectively extract multiscale features, inspired by UperNet \cite{2018UperNet} and Pyramid Pooling Module (PPM) \cite{zhao2017pyramid}, we stacked the building blocks constructed with this attention mechanism into a symmetric U-shaped encoder-decoder structure, in which the feature maps of different resolutions from the decoder component are fused to yield better segmentation results. Finally, we conduct experiments on the Intrapartum Ultrasound Grand Challenge 2024 dataset and verify the superiority of our method.

The remainder of this paper is organized as follows: Section \ref{method} provides a detailed description of the proposed methods. Section \ref{experiments} reports and analyzes the experimental setup and results. Section \ref{discussion} provides discussion of the experimental results. Section \ref{conclusion} presents our conclusion.

\section{Method}
\label{method}
In this section, we will first introduce the attention mechanism used in DSSAU-Net: Dual Sparse Selection Attention (DSSA). Next, we will provide a detailed description of the core module, the DSSA block, which is built upon this attention mechanism. Finally, we will outline the overall architecture of the network.
\begin{figure*}
\centering
\includegraphics[width=\textwidth]{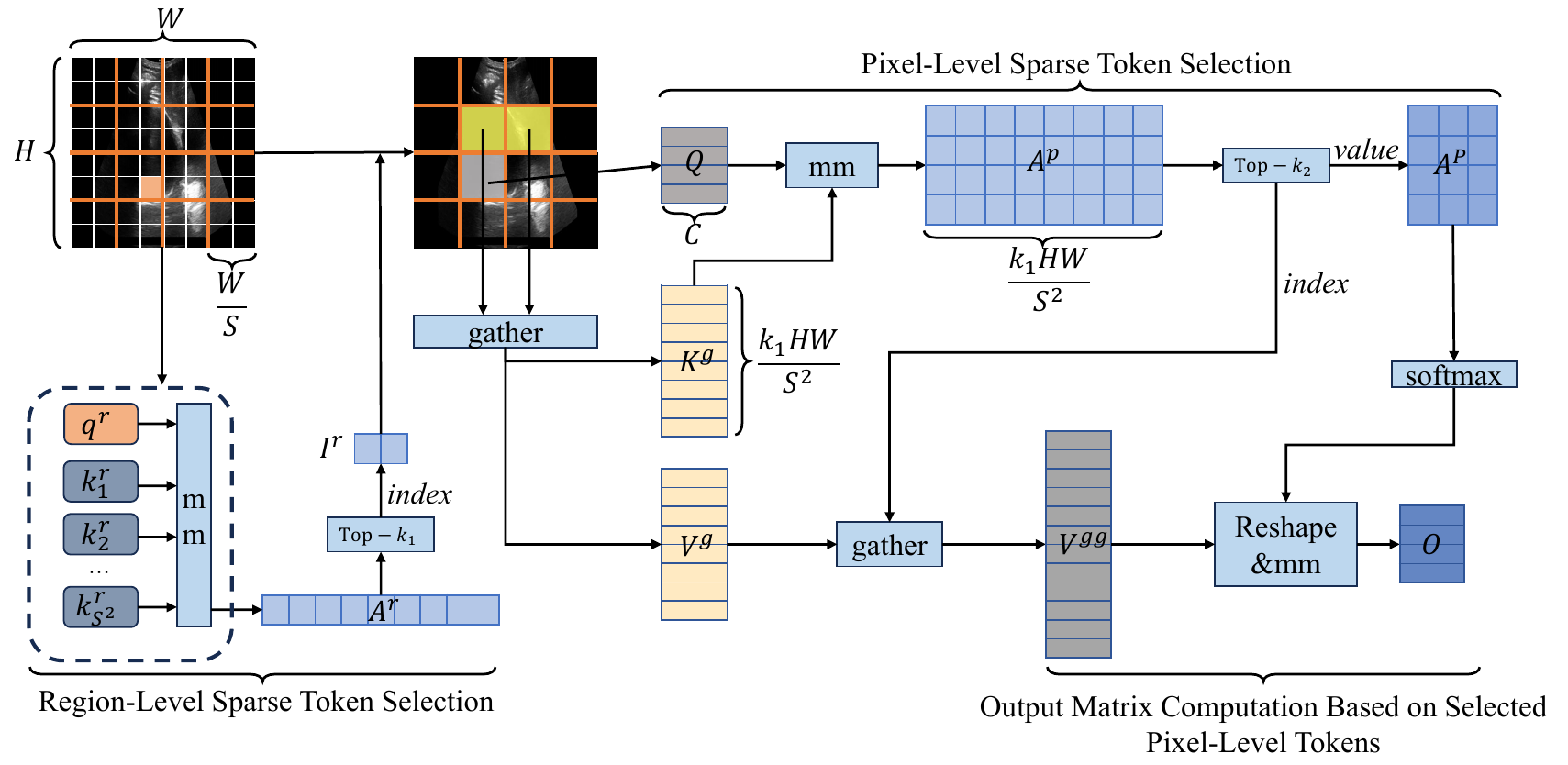}
\caption{Illustration of dual sparse selection attention (DSSA). DSSA performs twice sparse token selections at both the region and pixel levels, which can reduce computational complexity while extracting the most relevant features.}
\label{block}
\end{figure*}

\subsection{Dual Sparse Selection Attention (DSSA)}
The core idea of DSSA is to explicitly perform sparse token selections at both region and pixel levels. It involves three main steps. First, for a given region-level token where the pixel-level token is located, mostly relevant regions are selected and irrelevant ones are filtered out according to the scores of attention matrix $\textbf{A}^r$ between region-level query $\mathbf{Q}^r$ and key $\mathbf{K}^r$. Second, for a given pixel-level token, mostly relevant pixels are selected, and irrelevant ones are filtered out by the scores of attention matrix  $\textbf{A}^p$ between pixel-level query $\mathbf{Q}^p$ and key $\mathbf{K}^p$. Finally, the output matrix $O$ is computed by matrix multiplication between the normalized attention matrix $\textbf{A}^p$ and the pixel-level value $\mathbf{V}^p$. The conceptual diagram of DSSA is shown in Fig. \ref{block}. Following BiFormer \cite{Zhu2023biformer}, for a given input image $\mathbf{X} $ of size $H \times W \times C$, we divide the image into $S \times S$ non-overlapped regions, and obtain the region image $\mathbf{X^r} \in \mathbb{R}^{S^{2}\times\frac{HW}{S^{2}}\times C}$. Then, the region tokens: query, key, value, $\textbf{Q}, \textbf{K}, \textbf{V}\in \mathbb{R}^{S^{2}\times\frac{HW}{S^{2}}\times C}$, can be derived with linear projections:
\begin{equation}
\mathbf{Q}=\mathbf{X^r}\mathbf{W}^q,\mathbf{K}=\mathbf{X^r}\mathbf{W}^k,\mathbf{V}=\mathbf{X^r}\mathbf{W}^v.
\end{equation}
where $S$ presents the number of regions, $r$ indicates that a token is region-level, ${\textbf{W}^q}, {\textbf{W}^k}, {\textbf{W}^v} \in{{\mathbb R}^{C \times C}}$ are corresponding projection weight matrices for the query, key, value, respectively.

As for the region-level sparse token selection, each region-level token query  $\textbf{Q}^r\in \mathbb{R}^{S^{2} \times C}$ and key $\textbf{K}^r\in \mathbb{R}^{S^{2} \times C}$ can be derived by averaging all the pixel-level tokens on the region token $\mathbf{Q}$ and key $\mathbf{K}$. The attention map $\textbf{A}^r \in \mathbb{R}^{S^{2} \times S^{2}}$ between region-level queries $\mathbf{Q^r}$ and keys $\mathbf{K^r}$ is obtained through matrix multiplication, representing the semantic relevance between regions. Then, we employ a top-$k_1$ operation to select the $k_1$ most relevant regions for each region containing a given pixel-level token and record their indices in $\textbf{I}^r \in \mathbb{N}^{S^{2} \times k_1}$. The process is described as follows:
\begin{equation}\textbf{A}^r=\textbf{Q}^r(\textbf{K}^r)^{\top},\end{equation}
\begin{equation}\textbf{I}^r=\mathrm{topkIndex}(\textbf{A}^r).\end{equation}
To fully leverage the GPU's acceleration capabilities, the filtered region-level tokens need to be gathered back into a matrix. 
\begin{equation}
    \textbf{K}^{g}=\mathrm{gather}(\mathbf{K},\textbf{I}^r),\textbf{V}^{g}=\mathrm{gather}(\mathbf{V},\textbf{I}^r),
\end{equation}
where ${\textbf{K}^g}, {\textbf{V}^g} \in {{\mathbb R}^{{S^2} \times {\frac{{k_1}HW}{S^2}} \times C}}$ are gathered key and value tensors, which will be further used to the pixel-level sparse token selection.

Concerning the pixel-level sparse token selection, for any pixel-level query in $\textbf{Q}$, we compute the relevance (i.e., attention matrix) between this query and the gathered pixel-level key in $\textbf{K}^g$, 
indicated as $\textbf{A}^{p}\in \mathbb{R}^{S^2 \times \frac{HW}{S^2} \times \frac{k_{1}HW}{S^2}}$ as follows:
% \highlightmod{ denoted as $\textbf{A}^{p}\in \mathbb{R}^{S^2 \times \frac{HW}{S^2} \times \frac{k_{1}HW}{S^2}}$ as follows}:
\begin{equation}\textbf{A}^p=\textbf{Q}(\textbf{K}^g)^{\top},\end{equation}
where $p$ means that this token is pixel-level. Additionally, since the first round of region-level sparse token selection involves an averaging operation within a region, which may retain noise features due to the overall high relevance. This could negatively impact the model's ability to effectively extract features. Therefore, in $\textbf{A}^{p}$, we perform a second round of pixel-level sparse token selection using a top-$k_2$ operation, which not only selects the $k_2$ most relevant tokens but implicitly removes noises. The values and indices corresponding to the selected tokens are stored and denoted as $\textbf{A}^{P}\in \mathbb{R}^{S^2 \times \frac{HW}{S^2} \times k_2}$ and $\textbf{I}^{P}\in \mathbb{N}^{S^2 \times \frac{HW}{S^2} \times k_2}$, respectively:
% \highlightmod{denoted as $\textbf{A}^{P}\in \mathbb{R}^{S^2 \times \frac{HW}{S^2} \times k_2}$ and $\textbf{I}^{P}\in \mathbb{N}^{S^2 \times \frac{HW}{S^2} \times k_2}$}, respectively:
\begin{equation}
    \textbf{I}^{P}=\mathrm{topkIndex}(\textbf{A}^{p}),
\end{equation}
\begin{equation}
    \textbf{A}^{P}=\mathrm{topkValue}(\textbf{A}^{p}),
\end{equation}
where $k_2$ is determined by the scaling factor $\lambda$:
\begin{equation}
    k_2=\lambda\frac{k_{1}HW}{S^{2}}.
\end{equation}

Similarly, to leverage GPU acceleration, $\textbf{V}^g$ is gathered based on $\textbf{I}^P$ to form $\textbf{V}^{gg}\in R^{S^{2}\times\frac{k_{1}HW}{S^{2}}\times k_{2}\times C}$: 
\begin{equation}
    \textbf{V}^{gg}=\mathrm{gather}(\textbf{V}^g,\textbf{I}^P).
\end{equation}
Finally, $\textbf{V}^{gg}$ is weighted with $\textbf{A}^P$. Additionally, to preserve more fine-grained information beneficial for pixel-level segmentation, we add a local context enhancement term ($\mathrm{LCE}(\textbf{V})$) to the final output $\mathbf{O}$, which is a $5 \times 5$ depth-wise convolution: 
\begin{equation}\textbf{O}=\operatorname{Attention}(\textbf{A}^{P}, \textbf{V}^{gg})+\mathrm{LCE}(\textbf{V}).\end{equation}

\subsection{DSSA Block}
\begin{figure*}
\centering
\includegraphics[width=\textwidth]{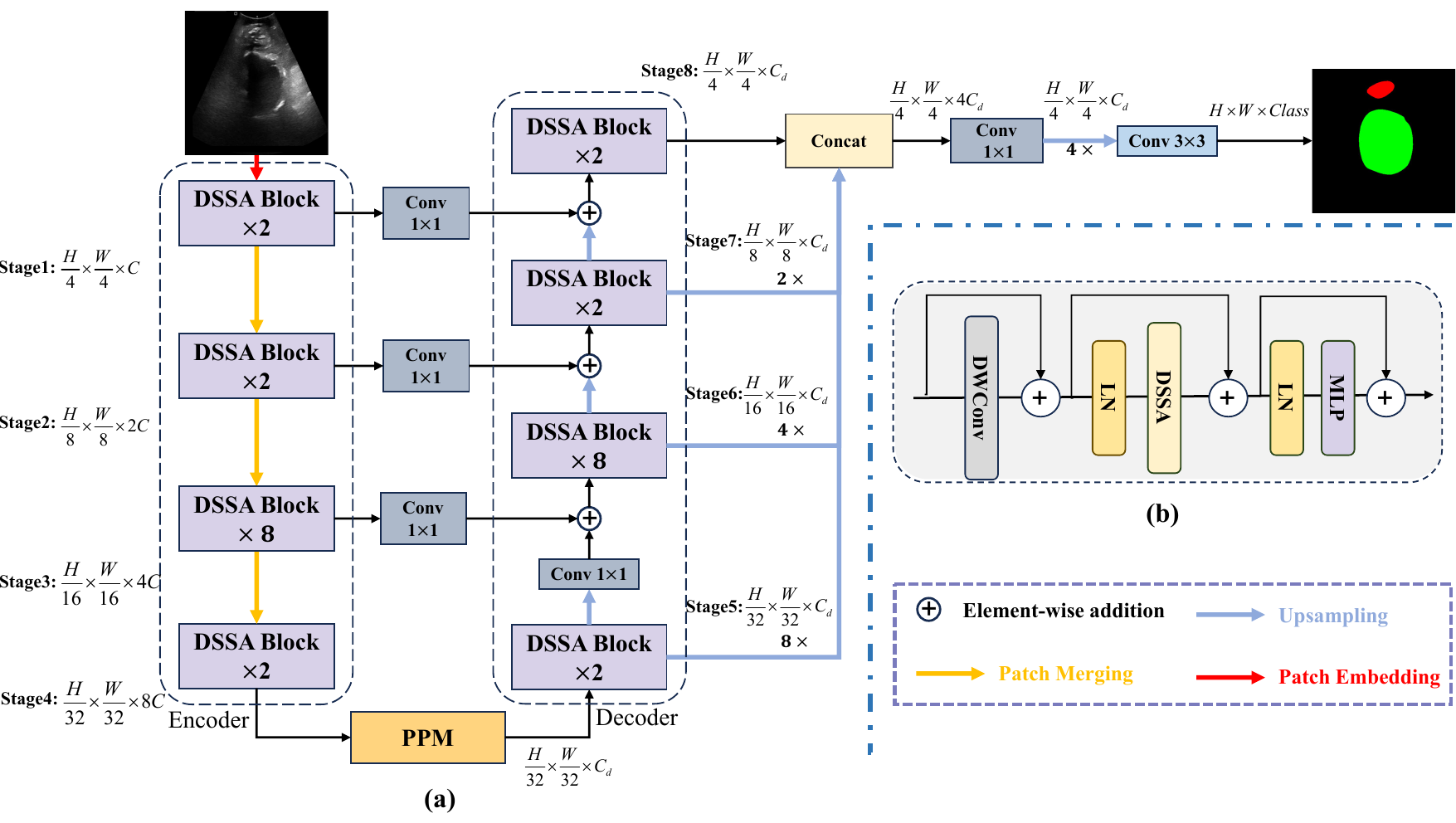}
\caption{(a): The overall architecture of DSSAU-Net, which is a u-shaped hybrid network and uses a sparse attention mechanism: dual sparse selection attention (DSSA) as a core building idea to hierarchically design the encoder-decoder structure. Furthermore, a pyramid pooling module (PPM) is used to fuse multiscale features, which is beneficial to improve the segmentation performance. (b): The details of the DSSA block.}
\label{architecture}
\end{figure*}

Based on this novel concept of DSSA attention, we can construct the core block of the model, the DSSA block, as shown in Fig. \ref{architecture}(b). Specifically, a $3 \times 3$ depth-wise convolution is first used to encode relative positional information, followed by a layer normalization. DSSA is then applied to compute attention, followed by another a layer normalization,  and the result is fed into a two-layer Multilayer Perceptron (MLP) module. During this process, three residual connections are employed to help alleviate the vanishing gradient and enhance the stability of the model. The DSSA block can be formulated as:
\begin{equation}
    \hat{\mathbf{z}} = \mathrm{DSSA}(\mathrm{LN}(\mathrm{DWConv}(\mathbf{z}^{l-1}) + \mathbf{z}^{l-1}) + \mathrm{DWConv}(\mathbf{z}^{l-1}) + \mathbf{z}^{l-1},
\end{equation}
\begin{equation}
    \mathbf{z}^{l} = \mathrm{MLP}(\mathrm{LN}(\hat{\mathbf{z}})) + \hat{\mathbf{z}},
\end{equation}
where $\mathbf{z}$ represents the output of each block, while $\hat{\mathbf{z}}$ indicates the output of the internal modules. $l$ denotes the $i$-th block.

\subsection{DSSAU-Net Architecture}
DSSAU-Net is a hybrid CNN-Transformer architecture with the encoder-decoder structure built upon DSSA blocks, as shown in Fig. \ref{architecture}(a). As for Stage1, it consists of an overlapped patch embedding layer with two 3×3 convolutions and two DSSA blocks, which is used to transform the input image $\mathbf{X}$ of sizes $H \times W \times 3$ into the resulting feature map $\mathbf{X}_1$ of size $\frac{H}{4} \times \frac{W}{4} \times C$: 
\begin{equation}
    \textbf{X}_1=\mathrm{DSSA Block_{\times 2}}(\mathrm{embedding}(\mathbf{X})),
\end{equation}
in Stage2, a Patch Merging layer is applied to reduce the number of tokens, and a linear embedding layer is used to increase the dimension $C$ to $2C$. Thus, the size of the feature map $\mathbf{X}_2$ output by Stage2, also containing two DSSA blocks, is $\frac{H}{8} \times \frac{W}{8} \times 2C$:
\begin{equation}
    \textbf{X}_2=\mathrm{DSSA Block_{\times 2}}(\mathrm{Merging}(\textbf{X}_1)),
\end{equation}
after that, similar process is repeated twice, the size of the corresponding feature maps $\mathbf{X}_3$ and $\mathbf{X}_4$ generated by Stage3 and Stage4 is $\frac{H}{16} \times \frac{W}{16} \times 4C$ and $\frac{H}{32} \times \frac{W}{32} \times 8C$, respectively, which are formulated as: 
\begin{equation}
    \textbf{X}_3=\mathrm{DSSA Block_{\times 8}}(\mathrm{Merging}(\textbf{X}_2)),
\end{equation}
\begin{equation}
    \textbf{X}_4=\mathrm{DSSA Block_{\times 2}}(\mathrm{Merging}(\textbf{X}_3)).
\end{equation}
On the encoder side, the number of blocks in each stage is set to 2, 2, 8, and 2, respectively.

Since $\mathbf{X}_4$ aggregates contextual information from different stages, inspired by UperNet \cite{2018UperNet}, we use the Pyramid Pooling Module (PPM) \cite{zhao2017pyramid} to take full advantage of the rich semantic information contained in $\mathbf{X}_4$. On the decoder side, the dimension of the feature map is fixed to $C_d$. This setup helps reduce the number of parameters and ensures that features at different scales are equally important. Subsequently, the feature map of size $\frac{H}{32} \times \frac{W}{32} \times {C_d}$ is fed into the hierarchically built decoder from Stage5 to Stage8. During this process, the feature map in each stage is processed with DSSA attention and fused with the feature map of the same resolution from the encoder. The resulting feature map is then upsampled to the output size of $\frac{H}{4} \times \frac{W}{4} \times {C_d}$. The overall process can be described as:
\begin{equation}
   \textbf{X}_5 = \mathrm{DSSA Block_{\times 2}}(\mathrm{PPM}(\textbf{X}_4)),
\end{equation}
\begin{equation}
   \textbf{X}_6 = \mathrm{DSSA Block_{\times 2}}(\mathrm{Conv}_{1 \times 1}(\textbf{X}_3) \oplus \mathrm{Conv}_{1 \times 1}(\mathrm{Up}_{2\times}(\textbf{X}_5))),
\end{equation}
\begin{equation}
    \textbf{X}_7 = \mathrm{DSSA Block_{\times 8}}(\mathrm{Conv}_{1 \times 1}(\textbf{X}_2) \oplus \mathrm{Up}_{2\times}(\textbf{X}_6)),
\end{equation}
\begin{equation}
    \textbf{X}_8 = \mathrm{DSSA Block_{\times 2}}(\mathrm{Conv}_{1 \times 1}(\textbf{X}_1) \oplus \mathrm{Up}_{2\times}(\textbf{X}_7)),
\end{equation}
where $\oplus$ presents element-wise addition.

Additionally, to effectively fuse the multiscale features in different stages, $\mathbf{X}_5$, $\mathbf{X}_6$, and $\mathbf{X}_7$ are upsampled to the same resolution as $\mathbf{X}_8$. All these upsampled feature maps are then concatenated along the channel dimension. After that, a $1 \times 1 $ convolution layer is applied, followed by $4\times$ upsampling, and another $3 \times 3 $ convolution layer. Finally, a feature map of resolution $H \times W \times Class$ is output to predict pixel-level segmentation. The process can be formulated as:
\begin{equation}
    \mathrm{Cat}=\mathrm{Concat}(\mathrm{Up}_{8\times}(\textbf{X}_5),\mathrm{Up}_{4\times}(\textbf{X}_6),\mathrm{Up}_{2\times}(\textbf{X}_7),\textbf{X}_8),
\end{equation}
\begin{equation}
    \mathrm{Output}=\mathrm{Conv}_{3 \times 3}(\mathrm{Up}_{4\times}(\mathrm{Conv}_{1 \times 1}(\mathrm{Cat}))).
\end{equation}

\subsection{Loss Function}
We adopt a hybrid loss function to train DSSAU-Net, combining dice loss ($\mathcal{L}_{dice}$) and cross-entropy loss ($\mathcal{L}_{ce}$). The dice loss helps alleviate class imbalance problems, while the cross-entropy loss ensures accurate pixel classification. These losses are defined as follows:
\begin{equation}
    \mathcal{L}_{dice}=1-\frac{2\sum_{i=1}^Np_ig_i}{\sum_{i=1}^Np_i+\sum_{i=1}^Ng_i},
\end{equation}
\begin{equation}
    \mathcal{L}_{ce}=-\frac{1}{N}\sum_{i=1}^N\left[g_i\log(p_i)+(1-g_i)\log(1-p_i)\right],
\end{equation}
\begin{equation}
    \mathcal{L}=\frac{1}{2}(\mathcal{L}_{dice}+\mathcal{L}_{ce}),
\end{equation}
where $p_i$ is the predicted probability of the $i$-th pixel, and $g_i$ is the ground truth of the $i$-th pixel.

\section{Experiments and Results}
\label{experiments}
\subsection{Datasets}
\label{datasets}
We conduct experiments on the competition dataset. The original dataset consists of 2,575 training images and 40 validation images. During training, all images are resized to $256 \times 256$.
\subsection{Metrics}
In this work, we employ three primary evaluation metrics to measure the segmentation performance of the model, namely the Dice Similarity Coefficient (DSC), the Hausdorff Distance (HD), and the Average Surface Distance (ASD). The specific calculations are presented as follows:
\begin{equation}
    \mathrm{DSC}(A,B)=\frac{2|A\cap B|}{|A|+|B|},
\end{equation}
\begin{equation}
    \mathrm{HD}(A, B) = \max \left( \max_{a \in A} \left( \min_{b \in B} \|a - b\| \right), \max_{b \in B} \left( \min_{a \in A} \|b - a\| \right) \right),
\end{equation}
\begin{equation}
    \mathrm{ASD}(A,B)=\frac{1}{2}\left(\frac{1}{|A|} \sum_{a \in |A|} \min_{b \in |B|} d(a,b)+\frac{1}{|B|} \sum_{b \in |B|} \min_{a \in |A|} d(b,a)\right).
\end{equation}

Additionally, to further demonstrate the effectiveness of our method in the ultrasound-assisted delivery process, during the final testing phase, we develop an automated program to calculate the Angle of Progression (AoP) and the Head-Symphysis Distance (HSD).

\subsection{Experimental Settings}
This work is conducted on a Geforce RTX 3090 GPU with 24GB of memory.
We meticulously design the experimental details. Specifically, the input images
are resized to a resolution of 256$\times$256, and the number of regions in each DSSA
stage is set to 8$\times$8. In the encoder, the number of channels in each stage
is set to [96, 192, 384, 768], and the $C_d$ of decoder is set to 64. Additionally, the initial learning rate is set to
1e-4. To improve the model’s generalization ability, we introduce several data
augmentation strategies, such as rotation, flipping, and contrast adjustment. The backbone network uses weights pre-trained on
the ImageNet dataset. The scaling factor $\lambda$ is set to 1/8.

\subsection{Results}
We train our DSSAU-Net on the training set and evaluate the performance of DSSAU-Net on the validation set mentioned in Subsection \ref{datasets}. The evaluation is conducted using three segmentation metrics: DSC, HD, and ASD, as well as two key parameters related to ultrasound-assisted diagnosis: AoP and HSD. The results are presented in Table \ref{tab1}. It can be seen that DSSAU-Net achieves excellent segmentation results, with DSC, HD, and ASD reaching up to 86.43, 31.08, and 8.39, respectively, and has a low number of parameters and FLOPs. This reflects the effectiveness of both the designed U-shaped hierarchical network for the pubic symphysis and fetal head segmentation task and the DSSA mechanism in reducing computational resource consumption.

To further show the performance of DSSAU-Net, we compare it with the methods used by other teams in the MICCAI IUGC 2024 competition. The performance of classification and segmentation on the \textit{test set} provided by the organizer, along with the final rank, is shown in Table \ref{tab2}. Our result wins fourth place in the classification and segmentation tasks in the competition. It can be seen that although DSSAU-Net did not achieve the highest final overall ranking, its performance on the segmentation task is still commendable. DSSAU-Net ranks second on both the HD and ASD metrics and third on the DSC metric, achieving an overall second place in the segmentation task. This demonstrates the strong capability of DSSAU-Net in handling segmentation tasks. However, its performance on the classification task is suboptimal, but this is in line with our expectations since our main goal is to develop a more accurate segmentation network. Overall, although DSSAU-Net performs well on the competition task and wins fourth place, it still has a large gap compared to the top-ranked team’s method. These gaps motivate us to continue our efforts to refine the network architecture and improve feature fusion strategies in future work.

\begin{figure*}
\centering
\includegraphics[width=\textwidth]{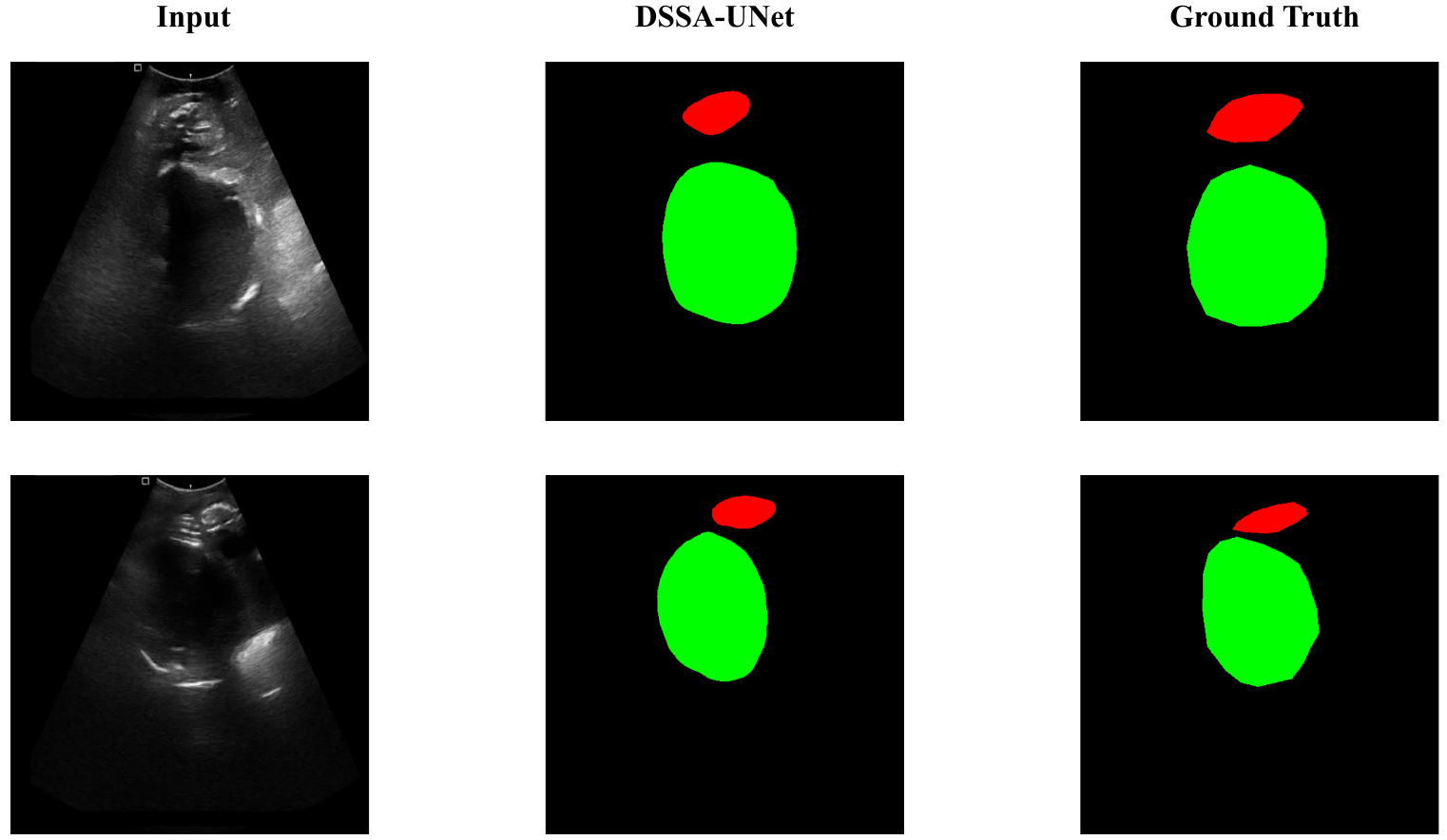}
\caption{The visualization results of DSSAU-Net on the Intrapartum Ultrasound Grand Challenge 2024 validation set. The red and green indicate the segmentation results of PS and FH, respectively.} 
\label{visual}
\end{figure*}

Additionally, we visualize the segmentation results and compare them with the ground truth, as shown in Fig. \ref{visual}. One can see that the segmented masks generated by our approach closely match the boundaries and shape of ground truth.

\begin{table}[htbp]
  \centering
  \caption{The segmentation performance of DSSAU-Net on the Intrapartum Ultrasound Grand Challenge 2024 validation set with 40 images provided by the organizer. The FLOPs are calculated with 256$\times$256 input.}
    \begin{tabular}{l|cc|ccc|cc}
        \hline
        % Methods &  AoP ($^{\circ}$) & HSD (\highlightmod{mm}) & DSC (\%) & HD (mm)& ASD (mm)& Params (M) & FLOPs (G)\\
        Methods &  AoP($^{\circ}$) & HSD(mm) & DSC(\%) & HD(mm)& ASD(mm)& Params(M) & FLOPs(G)\\
        \hline
        DSSAU-Net & 9.88 & 10.52 & 86.34 & 31.08 & 8.39 & 29.25 & 7.15\\
        \hline
    \end{tabular}%
  \label{tab1}%
\end{table}%

\begin{table}[htbp]
  \centering
  \caption{The performance of classification and segmentation on the \textit{test set} provided by the organizer. Our group, CQUT-Smart, adopts the DSSAU-Net and achieves a final rank of fourth. The symbol $\uparrow$ indicates the larger the better. The symbol $\downarrow$ signifies the smaller the better.}
    \begin{tabular}{l|cccc|ccc|cc|c}
        \hline
        \multirow{2}{*}{Group} & \multicolumn{4}{c|}{Classification} & \multicolumn{3}{c|}{Segmentation} & \multicolumn{2}{c|}{Biometry} & \multirow{2}{*}{Rank}\\
         \cline{2-10}
        & Acc $\uparrow$ & F1 $\uparrow$ & AUC $\uparrow$ & MCC $\uparrow$  & DSC $\uparrow$ & ASD $\downarrow$ & HD $\downarrow$ & AoP  $\downarrow$ & HSD  $\downarrow$ &  \\
        \hline
        ganjie & 74.41 & 75.55 & 78.02 & 36.48 & 84.75 & 13.00 & 38.72 & 10.43 & 11.45 & 1\\
        vicbic & 56.70 & 63.06 & 68.68 & 12.86 & 88.57 & 9.43 & 28.42 & 9.49 & 10.39 & 2\\
        BioMedIA & 66.19 & 54.22 & 71.26 & 21.45 & 86.33 & 12.22 & 41.11 & 9.16 & 11.68 & 3\\
        CQUT-Smart & 63.19 & 67.80 & 62.25 & 24.23 & 85.35 & 11.07 & 37.38 & 10.75 & 10.79 & 4\\
        nkdinsdale95 & 67.89 & 74.88 & 75.71 & 25.19 & 81.69 & 14.39 & 40.44 & 15.33 & 15.23 & 5\\
        baseline & 47.99 & 45.15 & 49.40 & 7.58 & 78.68 & 22.65 & 89.10 & 13.20 & 19.72 & 6\\
        hhl hotpot & 57.07 & 69.30 & 43.83 & 0 & 47.67 & 82.34 & 242.07 & 62.45 & 58.06 & 7\\
        serikbay & 42.93 & 0 & 44.10 & 0 & 72.63 & 56.21 & 134.79 & 40.14 & 89.21 & 8\\
        \hline
    \end{tabular}%
  \label{tab2}%
\end{table}%

\subsection{Ablation Study}
In this section, to deeply explore the specific impact of each component of the DSSAU-Net on overall performance, we follow the above experimental settings and conduct a series of ablation studies on the provided training set and validation set. Specifically, we systematically analyze the effects of different factors, including the number of skip connections, the multiscale feature fusion module, and the selection of key hyperparameters in the proposed Dual Sparse Selection Attention.

\subsubsection{Effect of the Number of Skip Connections:} Skip connection helps compensate for information loss during downsampling, which has been proved by previous studies\cite{ronneberger2015u,cao2022swin}. However, different network structures have varying complexities, and an excessive number of skip connections may not only be detrimental to segmentation performance but also increase the complexity of the network. Therefore, we conduct ablation experiments at resolutions of 1/4, 1/8, and 1/16 to explore the impact of different numbers of skip connections on the performance of DSSAU-Net. The results are shown in Table \ref{tab3}. When the skip connection is not used, the model has the worst performance on all segmentation metrics and biometry parameters. As the number of skip connections increases, the segmentation performance of the model gradually improves. The best performance is achieved when all skip connections are utilized at three resolutions. This is because multiscale skip connections are beneficial for compensating for the spatial information loss caused by downsampling and integrating high-level semantic information from deeper layers into the corresponding decoder layers, thereby enhancing segmentation performance. Thus, we report our final results conditioned on using all skip connections.
\begin{table}[htbp]
\centering
\caption{Ablation study of the number of skip connections.}
% \resizebox{1.0\textwidth}{!}{
% }
\begin{tabular}{c|cccc|cc|ccc}
\hline
\multirow{2}{*}{\# Skip Connection} &\multicolumn{4}{c|}{Connection Place} & \multirow{2}{*}{AoP $\downarrow$}  & \multirow{2}{*}{HSD $\downarrow$} & \multirow{2}{*}{DSC $\uparrow$}  & \multirow{2}{*}{HD $\downarrow$} & \multirow{2}{*}{ASD $\downarrow$} \\
\cline{2-5}
&no skip& 1/4&1/8&1/16&&&&&\\
\hline
0 &  $\checkmark$ & & & & 10.18 & 12.14 & 85.84 & 32.90 & 9.23\\
1 &   & $\checkmark$& & & 10.05 & 12.42 & 86.03 & \underline{31.55} & 9.15\\
2 &   &$\checkmark$ &$\checkmark$ & & \underline{9.98} & \underline{11.79} & \underline{86.11} & 31.57 & \underline{8.89}\\
3 &  & $\checkmark$ & $\checkmark$&$\checkmark$ & \textbf{9.88} & \textbf{10.52} & \textbf{86.34} & \textbf{31.08} & \textbf{8.39}\\
\hline
\end{tabular}
\label{tab3}
\end{table}

\subsubsection{Effect of Multiscale Feature Fusion Module:} The multiscale feature fusion (MFF) module is the core component of DSSAU-Net used to achieve accurate segmentation. Thus, we design the ablation experiment to analyze its effectiveness. The results are presented in Table \ref{tab4}. It is evident that compared to DSSAU-Net without the MFF module, the performance of DSSAU-Net with the MFF module improves by 0.37, 1.2, and 0.81 in terms of DSC, HD, and ASD, respectively. Consistent improvement can also be observed with respect to AoP and HSD. We believe that the reason may be that the shallow feature maps have higher resolution and contain more local spatial information, while the deep feature maps have lower resolution and cover rich global semantic information. Both of these advantages can be leveraged by the MFF module, thereby achieving more accurate segmentation.
\begin{table}[htbp]
  \renewcommand{\arraystretch}{1.25}
  \centering
  \caption{Ablation study of multiscale feature fusion (MFF) module.}
    \begin{tabular}{l|cc|ccc}
        \hline
        Method &  AoP $\downarrow$ & HSD $\downarrow$ & DSC $\uparrow$ & HD $\downarrow$ & ASD $\downarrow$\\
        \hline
        DSSAU-Net(w/o MFF) & 10.39 & 12.53 & 85.97 & 32.28 & 9.20\\
        DSSAU-Net & \textbf{9.88} & \textbf{10.52} & \textbf{86.34} & \textbf{31.08} & \textbf{8.39}\\
        \hline
    \end{tabular}
  \label{tab4}
\end{table}

\subsubsection{Effect of the Number of Top-$k$ Tokens:} 
The dual sparse selection mechanism in DSSA not only significantly reduces computational complexity but also enables the extraction of accurate features. However, selecting different numbers of region-level tokens and pixel-level tokens may impact the performance of DSSAU-Net. Therefore, we design the ablation study to choose appropriate $k_1$ and $\lambda$ values for the region-level tokens and pixel-level tokens. For a fair comparison, this experiment does not use pre-trained weights. The results are presented in Table \ref{tab5}. One can see that when $k_1$ is set to [1, 4, 16, 64], a better segmentation performance is achieved compared to [2, 8, 32, 64]. We analyze that this is because selecting too many regions may introduce more noise tokens, which is detrimental to learning effective features. Additionally, we fix $k_1$ to [1, 4, 16, 64] and further study the effect of the values of $\lambda$. It can be seen that when setting $\lambda$ to 1/8, the model can achieve the best performance. There is a similar observation when $k_1$ is set to [2, 8, 32, 64] and $\lambda$ is set to 1/8, the model also performs better compared to other $\lambda$ values. This is easily understandable because when fewer tokens are selected during the pixel-level sparse selection, the model can learn fewer features. Conversely, selecting more tokens introduces more noise. Moreover, it can be seen from Table \ref{tab5} that the FLOPs vary significantly with different values of $\lambda$, demonstrating the advantage of DSSA in reducing computational complexity. Therefore, we chose a compromise value of $\lambda$ as the hyperparameter of the model.
\begin{table}[htbp]
  \centering
  \caption{Ablation study of the number of top-$k$ tokens.}
    \begin{tabular}{c|c|cc|ccc|c}
        \hline
        $k_1$& $\lambda$ &  AoP $\downarrow$ & HSD $\downarrow$ & DSC $\uparrow$ & HD $\downarrow$ & ASD $\downarrow$ & FLOPs (G)\\
        \hline
        \multirow{3}{*}{1,4,16,64} & 1/4 & \underline{13.79} & \underline{16.94} & 82.61 & \underline{40.16} & 10.84 & 7.17\\
        & 1/8 & \textbf{13.50} & \textbf{16.89} & \textbf{83.42} & \textbf{38.15} & \textbf{10.41} & 7.15\\
        & 1/16 & 15.24 & 17.54 & 82.68 & 41.62 & 10.93 & 7.14\\
        \hline
        \multirow{3}{*}{2,8,32,64} & 1/4 & 15.57 & 17.17 & 82.61 & 41.60 & 11.01 & 7.39\\
        & 1/8 & 15.54 & 17.24 & \underline{82.73} & 40.49 & \underline{10.79} & 7.35\\
        & 1/16 & 15.84 & 19.16 & 82.72 & 42.52 & 29.25 & 7.32\\
        \hline
    \end{tabular}
  \label{tab5}
\end{table}

\section{Discussion}
\label{discussion}
In this paper, we have validated the performance of DSSAU-Net in ultrasound-assisted delivery. However, From Fig. \ref{visual}, it can be seen that although DSSAU-Net has successfully segmented both PS (presentation of the fetal head) and FH (fetal head), closely aligning with the ground truth, there is still room for improvement in the precision of edge segmentation. For instance, segmenting the boundaries between target and background is challenging due to low contrast. Therefore, enhancing DSSAU-Net's capability to extract local information and improving its performance under low-contrast imaging conditions will be the focus of future work.

\section{Conclusion}
\label{conclusion}
In this work, we propose DSSAU-Net, a CNN-Transformer hybrid network for fetal head and pubic symphysis segmentation. The network is constructed by stacking efficient DSSA blocks, forming symmetrical encoder-decoder structure. Additionally, we introduce skip connections with convolution operations, and a pyramid pooling module to capture richer semantic information. The results on the challenge dataset demonstrate that our approach is capable of achieving accurate and efficient medical image segmentation. In future work, we will further explore more efficient attention mechanisms and investigate the fusion of feature information across different levels to enhance the model's ability for both global and local modeling.

\bibliographystyle{splncs04}
\bibliography{ref}

\begin{thebibliography}{10}
\providecommand{\url}[1]{\texttt{#1}}
\providecommand{\urlprefix}{URL }
\providecommand{\doi}[1]{https://doi.org/#1}

\bibitem{bellussi2017intrapartum}
Bellussi, F., Ghi, T., Youssef, A., Cataneo, I., Salsi, G., Simonazzi, G., Pilu, G.: Intrapartum ultrasound to differentiate flexion and deflexion in occipitoposterior rotation. Fetal Diagnosis and Therapy  \textbf{42}(4),  249--256 (2017)

\bibitem{boyle2013primary}
Boyle, A., Reddy, U.M., Landy, H.J., Huang, C.C., Driggers, R.W., Laughon, S.K.: Primary cesarean delivery in the united states. Obstetrics \& Gynecology  \textbf{122}(1),  33--40 (2013)

\bibitem{cai2023pubic}
Cai, P., Lu, J., Li, Y., Lan, L.: Pubic symphysis-fetal head segmentation using pure transformer with bi-level routing attention (2023), \url{https://arxiv.org/abs/2310.00289}

\bibitem{cao2022swin}
Cao, H., Wang, Y., Chen, J., Jiang, D., Zhang, X., Tian, Q., Wang, M.: Swin-unet: Unet-like pure transformer for medical image segmentation. In: European conference on computer vision. pp. 205--218. Springer (2022)

\bibitem{chen2024fetal}
Chen, Z., Lu, Y., Long, S., Campello, V.M., Bai, J., Lekadir, K.: Fetal head and pubic symphysis segmentation in intrapartum ultrasound image using a dual-path boundary-guided residual network. IEEE Journal of Biomedical and Health Informatics  (2024)

\bibitem{chen2024direction}
Chen, Z., Ou, Z., Lu, Y., Bai, J.: Direction-guided and multi-scale feature screening for fetal head--pubic symphysis segmentation and angle of progression calculation. Expert Systems with Applications  \textbf{245},  123096 (2024)

\bibitem{cohen2017prolonged}
Cohen, S., Lipschuetz, M., Yagel, S.: Is a prolonged second stage of labor too long? Ultrasound in Obstetrics \& Gynecology  \textbf{50}(4),  423--426 (2017)

\bibitem{dupuis2005fetal}
Dupuis, O., Ruimark, S., Corinne, D., Simone, T., Andr{\'e}, D., Ren{\'e}-Charles, R.: Fetal head position during the second stage of labor: comparison of digital vaginal examination and transabdominal ultrasonographic examination. European Journal of Obstetrics \& Gynecology and Reproductive Biology  \textbf{123}(2),  193--197 (2005)

\bibitem{fiorentino2023review}
Fiorentino, M.C., Villani, F.P., Di~Cosmo, M., Frontoni, E., Moccia, S.: A review on deep-learning algorithms for fetal ultrasound-image analysis. Medical image analysis  \textbf{83},  102629 (2023)

\bibitem{ghi2018isuog}
Ghi, T., Eggeb{\o}, T., Lees, C., Kalache, K., Rozenberg, P., Youssef, A., Salomon, L., Tutschek, B.: Isuog practice guidelines: intrapartum ultrasound. Ultrasound in Obstetrics \& Gynecology  \textbf{52}(1),  128--139 (2018)

\bibitem{malvasi2014occiput}
Malvasi, A., Tinelli, A., Barbera, A., Eggeb{\o}, T., Mynbaev, O., Bochicchio, M., Pacella, E., Di~Renzo, G.: Occiput posterior position diagnosis: vaginal examination or intrapartum sonography? a clinical review. The Journal of Maternal-Fetal \& Neonatal Medicine  \textbf{27}(5),  520--526 (2014)

\bibitem{ramphul2014instrumental}
Ramphul, M., Ooi, P.V., Burke, G., Kennelly, M.M., Said, S.A., Montgomery, A.A., Murphy, D.J.: Instrumental delivery and ultrasound: a multicentre randomised controlled trial of ultrasound assessment of the fetal head position versus standard care as an approach to prevent morbidity at instrumental delivery. BJOG: An International Journal of Obstetrics \& Gynaecology  \textbf{121}(8),  1029--1038 (2014)

\bibitem{ronneberger2015u}
Ronneberger, O., Fischer, P., Brox, T.: U-net: Convolutional networks for biomedical image segmentation. In: Medical image computing and computer-assisted intervention--MICCAI 2015: 18th international conference, Munich, Germany, October 5-9, 2015, proceedings, part III 18. pp. 234--241. Springer (2015)

\bibitem{challenge2013evaluation}
Rueda, S., Fathima, S., Knight, C.L., et~al.: Evaluation and comparison of current fetal ultrasound image segmentation methods for biometric measurements: A grand challenge. IEEE Transactions on Medical Imaging  \textbf{33}(4),  797--813 (2014)

\bibitem{2018UperNet}
Xiao, T., Liu, Y., Zhou, B., Jiang, Y., Sun, J.: Unified perceptual parsing for scene understanding. In: Ferrari, V., Hebert, M., Sminchisescu, C., Weiss, Y. (eds.) Computer Vision -- ECCV 2018. pp. 432--448. Springer International Publishing, Cham (2018)

\bibitem{Zhao2019ExplicitST}
Zhao, G., Lin, J., Zhang, Z., Ren, X., Su, Q., Sun, X.: Explicit sparse transformer: Concentrated attention through explicit selection (2019), \url{https://arxiv.org/abs/1912.11637}

\bibitem{zhao2017pyramid}
Zhao, H., Shi, J., Qi, X., Wang, X., Jia, J.: Pyramid scene parsing network. In: Proceedings of the IEEE conference on computer vision and pattern recognition. pp. 2881--2890 (2017)

\bibitem{Zhu2023biformer}
Zhu, L., Wang, X., Ke, Z., Zhang, W., Lau, R.: Biformer: Vision transformer with bi-level routing attention. In: 2023 IEEE/CVF Conference on Computer Vision and Pattern Recognition (CVPR). pp. 10323--10333 (2023)

\end{thebibliography}

\end{document}